\documentclass[preprint,12pt,number]{elsarticle}
\biboptions{square,sort,comma,numbers}

\usepackage{xcolor}

\usepackage{amssymb}
\usepackage{amsmath}
\usepackage{multirow}
\usepackage{booktabs}
\usepackage{setspace}
\usepackage{lmodern}
\usepackage{anyfontsize}

\journal{Nuclear Physics B}

\makeatother

\makeatletter
\def\thefootnote{}
\makeatother
\begin{document}

\begin{frontmatter}



\title{Hidden-Shot: Towards One-Shot Task Generalization for Low-Level Vision Generalist Models} %% Article title


%% Authors
\author[a]{Shao-Jun Xia\corref{cor1}}
\ead{shaojun.xia@duke.edu}

\author[b]{Xianzheng Ma}
% \ead{xianzheng@robots.ox.ac.uk}

\author[c]{Zichong Meng}
% \ead{meng.zic@northeastern.edu}

%% Corresponding author
\cortext[cor1]{Corresponding author.}

% \fntext[fn1]{This preprint is under submission.}

%% Affiliations
\affiliation[a]{
    organization={Pratt School of Engineering, Duke University},
    city={Durham},
    state={NC},
    postcode={27708},
    country={USA}
}

\affiliation[b]{
    organization={Department of Engineering Science, University of Oxford},
    city={Oxford},
    addressline={Parks Road},
    postcode={OX1 3PJ},
    country={UK}
}

\affiliation[c]{
    organization={Khoury College of Computer Science, Northeastern University},
    city={Boston},
    state={MA},
    postcode={02115},
    country={USA}
}

%% Abstract
\begin{abstract}
Despite the intense engagement surrounding low-level vision generalist models, their effectiveness in zero/few-shot scenarios beyond learned tasks remains unverified. The primary challenge of developing an ideal generalist lies in achieving the ability to generalize from new unseen tasks, which also can be assessed by matched quantitative criteria. Existing methods have made some progress in prompt engineering but have not systematically explored this gap across a wide range of low-level visual tasks. Stimulated by the problem, we propose Hidden-Shot, an implicit prompt mechanism aimed at exploring low-level task adaptation in a vision generalist model. Specifically, the method extracts implicit visual task-based information, utilizes a global task-aware textural prompt, and selectively merges implicit information with in-task processing information to enhance one-shot capabilities in new tasks. The overall design performs direct injection in a cost-effective manner, while minimally altering the architecture of the original generalist model. Additionally, we introduce a data-driven evaluation framework termed C/U assessment to cover two basic scenarios, 3C4U (3 conventional and 4 unconventional tasks) for retraining existing models and 3C7U (3 conventional and 7 unconventional tasks) for training from scratch, as a comprehensive assessment to systematically test the generalization ability of low-level generalist models. Experiments on seven and ten datasets outperform the state-of-the-art vision generalist model, respectively verified by 3C4U and 3C7U framework. Our presented Hidden-Shot approach demonstrates superior performance on one-shot new tasks while maintaining consistent performance on existing tasks.
\end{abstract}

%% Keywords

\end{frontmatter}

\begingroup
\renewcommand\thefootnote{}
\footnotetext{This preprint is under submission.}
\addtocounter{footnote}{-1}
\endgroup



\section{Introduction}
\label{sec:introduction}

Large vision generalist model \citep{unified-io, pix2seqv2} has a significant impact on unifying diverse vision tasks. By jointly training on various tasks, primarily through supervised or self-supervised ways such as masked image modeling \citep{vit-mae}, the model can perform well on multiple tasks. 
This ability mainly derives from two aspects: (1) Vision transformer lays a structural foundation for model to acquire stronger capacity. (2) The successful utilization of in-context learning (ICL) in natural language processing (NLP) for unifying tasks gives more experience for computer vision (CV) community.

The fundamental principle for generalist models in NLP, \textit{e.g.}, GPT\citep{gpt}, depends on the same input and output format. The language token serves as a natural and unified data format for all tasks. Differently, vision tasks have multiple output formats, ranging from class labels (classification), bounding boxes (detection) to semantic masks (semantic segmentation), and RGB images (image restoration), thus converting most tasks into a sequential completion is tough in the vision domain. Though the output formats of low-level tasks are images, unifying the most tasks is also not easy. From language perspective, it seems that many low-level tasks start with the same prefix ``de'' or ``re'', \textit{e.g.}, ``dehazing'', ``deblurring'', ``denoising'', ``deraining'', ``demoiring'' or ``relighting'', ``reflection removal'', the actual task domains are highly disparate from vision perspective. For example, dehazing and deraining seem to both process white noise in visual appearance. But essentially, dehazing focuses on deep information related to natural characteristics, while deraining aims to remove the additive noise in the image.

Except for unifying multiple tasks, most generalist models face more difficulties in effectively generalizing to new low-level tasks. In recent studies, MAE-VQGAN \citep{mae-vqgan} and Painter \citep{painter} are two representative models that deal with a few low-level tasks, \textit{e.g.}, denoising, deraining, and light enhancement. ProRes \citep{prores} incorporates degradation-aware visual prompts to handle new tasks, but the approach involves complex optimization processes accompanying the design of loss. PromptGIP \citep{promptgip} addresses new tasks through a specific mechanism of fine-tuning. Predictably, the method will bring a high computational cost of previously learned tasks. Subsequent works such as PromptIR \citep{promptir} and InstructIR \citep{instructir}, primarily include drastically modifying the structure of prompt networks or enhancing the prompt for current tasks. However, these two methods have not explored the full capability of adding new tasks. This also constitutes the gap hindering universal large-scale models from self-evolution.

One-shot generalization has become increasingly important for practical deployment \citep{yu2025learning} because low-level vision systems often require rapid adaptation to new environments with minimal task-specific data, particularly in scenarios where collecting or annotating large-scale paired degradation datasets is infeasible \citep{pereg2024one}. Previous studies \citep{chen2024learning, wang2024image} in low-level vision generalist models demonstrate that real-world degradations exhibit substantial variability, further necessitating the need for flexible and rapidly adaptable learning paradigms.
Current single-task low-level models \citep{hu2026revisiting, shi2025image} trained for each specific degradation (\textit{e.g.}, dehazing, deraining) usually struggle when confronted with unseen combinations or contexts, as they tend to over-fit to training conditions or require extensive retraining. Regrettably, retraining models for every new scenario imposes prohibitive computational and time costs. Recent advances in retraining paradigms \citep{liu2023degae} and unified text-induced processors \citep{duan2024uniprocessor} have significantly improved the generalization flexibility of low-level vision models. However, these approaches still suffer from limited task scalability or overly complex specialized architectures with high computational costs, which constrain their applicability to diverse and dynamic real-world settings.

Rethinking the generalization problem in the field, we naturally argue that a real low-level vision generalist model should basically satisfy one key-point: 
% \textit{The generalist model has the ability to generalize to new unseen tasks and adopt one-shot scenarios in a low-cost way.}
\textit{A low-level generalist model should be capable of self-enhancement by generalizing to new, unseen tasks in a cost-effective manner, such as through few-shot or even one-shot learning, without damaging its performance on previous seen tasks.}

In this paper, we construct an innovative lightweight prompt framework inserted into a typical generalist model, \textit{i.e.}, Hidden-Shot for low-level one-shot generalization. The framework takes into full consideration the implicit visual features and global-aware textual features from low-level CV tasks. By utilizing implicit representation learning, the method injects a task-aware and language-assisted prompt to the decoder part of a generalist model, which guides the model to generalize to unseen tasks without intricate network coupling. Our main contributions can be summarized into four parts:
\begin{itemize}
    \item We present a fine-grained prompt mechanism to improve the ability of unseen task generalization for generalist vision model, by generating and injecting low-cost implicit task-based prompts to the decoder of the original network.
    \item Our Hidden-Shot method utilizes a new strategy integrated with lightweight global vectors, which proves language information of low-level vision tasks can help to enhance the one-shot generalization ability for new unseen tasks.
    \item The proposed prompt-tuning approach achieves advanced performance in one-shot generalization setting and maintains stable performance on existing tasks, avoiding catastrophic forgetting.
    \item We provide a data-driven evaluation framework to systematically assess the few-shot generalization in low-level vision tasks as a first attempt, which will broadly contribute to this field.
\end{itemize}

\section{Related Work}
\label{sec:related_work}
\subsection{Vision Unified Modeling}
The concept of addressing diverse vision tasks with a unified architecture has been a longstanding focus in CV community. Driven by the progress in NLP, early works such as Unified-IO \citep{unified-io} and Pix2seqv2 \citep{pix2seqv2} handle diverse vision tasks as a sequence-to-sequence generation problem. These methods employ sequences of discrete tokens to represent input and output across diverse vision tasks. Recent works have shifted focus and started to use advanced vision models such as vision transformers and diffusion models for better performance. For example, InstructCV \citep{gan2024instructcv} and InstructDiffusion \citep{geng2024instructdiffusion} leverage the latent diffusion model to reform vision tasks into multimodal guided image generation tasks by denoising.

However, the approaches above often necessitate full retraining or fine-tuning to perform new tasks beyond trained tasks. This leads to issues such as increasing demand for training resources. In our work, we introduce an efficient Lightweight method that enables unified vision model (vision generalist model) for task generalization while reducing catastrophic forgetting.

\subsection{Visual In-Context Learning}
ICL has long been a heated research direction in NLP. The emergence of GPT-3 \citep{gpt} has showcased the ability to learn and adapt task context through language prompts. Recently, CV researchers have also made progress on visual in-context learning (VICL). For instance, VICL for image inpainting is first carried out by MAE-VQGAN \citep{mae-vqgan}. MAE-VQGAN proposes structuring input images with visual prompts. The mechanism consists of one image pair, a query image, and an empty image. The model then performs vision tasks by inferring the missing information. The way is later enhanced by Painter \citep{painter}, which uniformly processes 3 low-level and 5 high-level vision tasks.

With the rapid development of diffusion models, several diffusion-based architectures have been shown to be beneficial for VICL \citep{oorloff2025stable}. 
Prompt Diffusion \citep{promptdiffusion} achieves more expressive VICL by using a diffusion model guided by example–prompt pairs. InstructDiffusion \citep{geng2024instructdiffusion} is also commonly adopted as a baseline model for comparisons in this domain, by adapting diffusion models through instruction-driven conditioning. Since most VICL settings are based on one-shot scenarios, there are also several advanced one-shot studies. FaceShot \citep{gao2025faceshot} introduces a one-shot, training-free portrait animation framework that brings any character to life from a single image while preserving identity. Similarly, StyleShot \citep{styleshot} provides a one-shot style transfer approach that captures and applies diverse artistic styles from a single reference image without any retraining. While these two methods have not been directly employed in VICL, they nevertheless exhibit strong potential for future application.

In another line of research, many papers primarily focus on maintaining the architectures and exploring better prompt structure to enhance trained tasks. ProRes \citep{prores} presents degradation-aware visual prompts that better fit image restoration tasks. PromptGIP \citep{promptgip} proposes structuring prompts as question answering. PromptIR \citep{promptir} adapts various inbuilt prompt blocks to enhance the capabilities of general image restoration. CONDENSER~\cite{wang2025embracing} explores multi-prompt collaboration by condensing complementary information from multiple demonstrations into a unified prompt representation. 
X-Prompt~\cite{sun2025x} introduces an auto-regressive visual learning framework that improves task generalization through in-context prompting. 
PromptHub~\cite{luo2026prompthub} further advances multi-prompt VICL with locality-aware fusion, concentration, and alignment for more effective prompt integration. Instead of improving the generality of generalist models via extra explicit prompts, we implicitly exploit the potential of VICL to address the task generalization ability for the first time. 

\subsection{Task Generalization}
Across all machine learning fields, models trained on previously learned tasks pose serious challenges for unseen tasks. \citep{taskinit} is considered as the first vision work on cross-task generalization. Early representative works such as Taskonomy \citep{taskonomy} and TTNet \citep{ttnet} focus on the correlation between vision tasks. Both TTNet and Taskonomy believe vision tasks are intercorrelated and it is possible to train each different model on one task. Then task generalization is achieved by the correlation or affinity matrices.

Many researchers \citep{dwivedi2019representation} have also studied task relationships for task generalization. However, these methods fail to function as multitask-enabled vision generalist models. The procedure of training each model on one task prior to learning the task relationships is limited in few-shot task generalization settings.
VTAGML~\citep{vtagml} then presents a multitask learnable transformer with generalization potential. However, each decoder part of the model corresponds with only one task, hence the model still suffers in few-shot generalization to various vision tasks. Differently, we present a novel and cost-effective prompt injection to achieve one-shot task generalization in this work, without the need for pre-defining task correlations.

\section{Methodology}
\label{sec:method}
\subsection{Framework Overview}
\label{sec:3.1}
\begin{figure*}[tp]
\centering
\includegraphics[width=0.95\columnwidth]{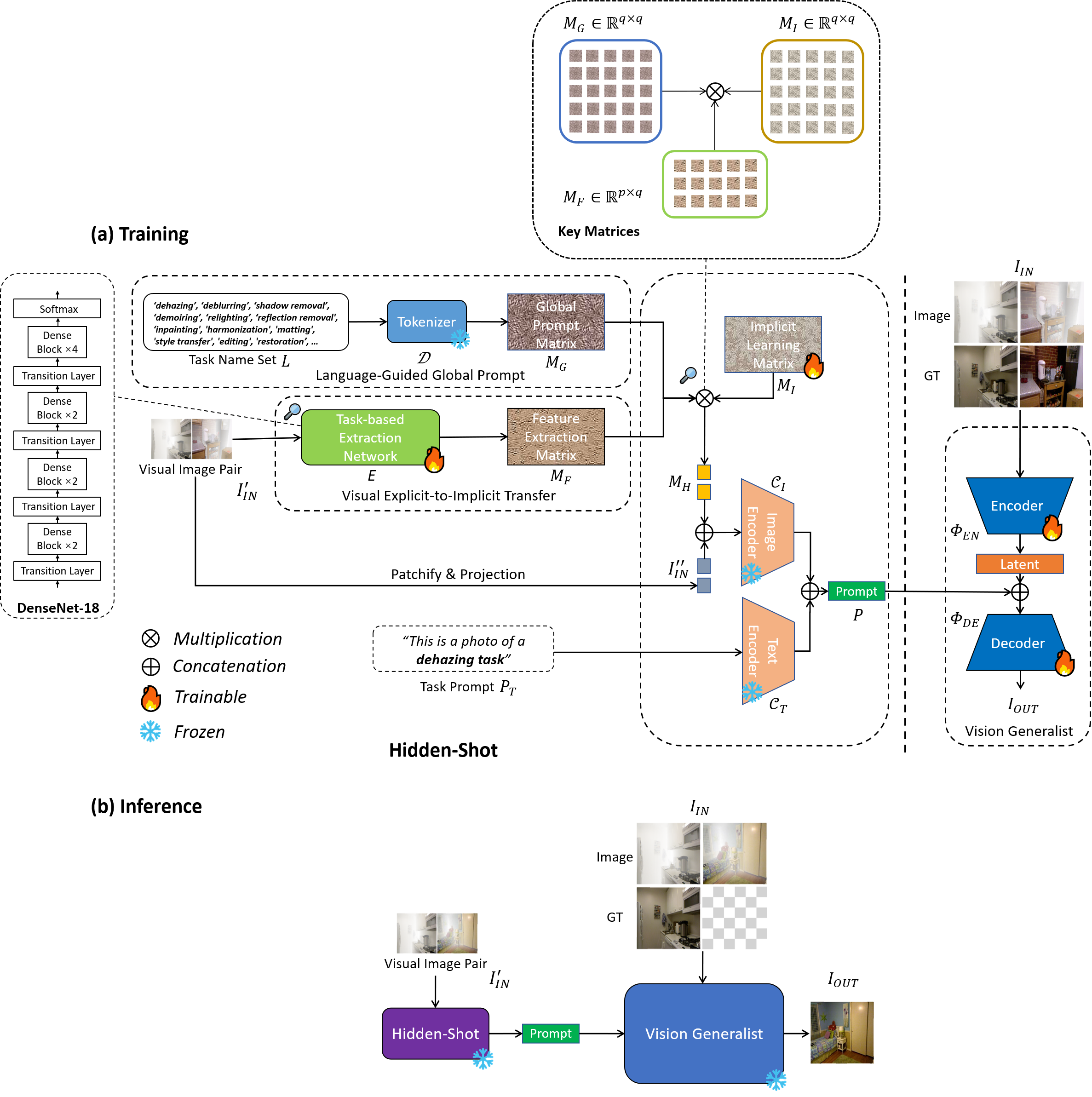}
\caption{Overview of Hidden-Shot framework. The right half illustrates the common structure of vision generalist models. The left half shows the prompt generation mechanism, including visual explicit-to-implicit transfer, language-guided global prompt, and implicit information combination.}
\label{fig:framework}
% \vspace{-2mm}
\end{figure*}

We present our designed pipeline solution in Figure
\ref{fig:framework}. The right half of the figure illustrates the process from input to output. Initially, the input $I_{IN} \in \mathbb{R}^{h \times w \times 3}$of visual prompted image pair (2 images and corresponding ground truth) undergoes an encoding phase via the Painter Encoder $\mathit\Phi_{EN}$, which extracts visual prompted latent embedding $LA$. Subsequently, the latent is concatenated with our generated prompt $P$. Inspired by the straightforward yet effective decoder-only feature injection from Mask2Former \citep{mask2former}, the mixed intermediate is fed into the Painter Decoder $\mathit\Phi_{DE}$ to get the final visual output $I_{OUT}$.

The left half of the figure depicts our prompt generation mechanism. Concurrent with $I_{IN}$, a visual image pair input $I_{IN}'$ (only 2 images) and a task-descriptive text prompt $P_T$ (\textit{e.g.}, ``This is a task of a [TASK NAME] task.'') are processed through our proposed Hidden-Shot. The architecture consists of three parts: a visual explicit-to-implicit transfer module, a language-guided global prompt module, and an implicit information combination module.

\textbf{Task-specific: Visual Explicit-to-Implicit Transfer Module} extracts visual features from $I_{IN}'$ through a task-based extraction network $\mathcal{D}$, and transforms these visual features into an implicit feature extraction matrix $M_{F}$ that encapsulates rich implicit visual task-based information.

\textbf{Task-agnostic: Language-guided global Prompt module} processes and extracts textual task information through a tokenizer $E$, and forms a global prompt matrix $M_{G}$ that contains implicit global textual task-aware features information.

\textbf{Integration: Implicit information combination module} utilizes a learnable implicit prompt matrix $M_{I}$ to selectively interact with feature extraction matrix $M_{F}$ and global prompt matrix $M_{G}$. Through matrix multiplication, the hidden product matrix $M_{H}$ is obtained, which encapsulates multimodal hidden task-based information.
$M_{H}$ is then concatenated with the patchified and projected $I_{IN}'$ to derive a composite CLIP image feature $M_{CI}$ (by CLIP image encoder $\mathcal{C}_{I}$). $M_{CI}$ is concatenated with the CLIP text feature $M_{CT}$ (by CLIP text encoder $\mathcal{C}_{T}$) to produce the coupled prompt $P$. The module integrates multimodal task-based and in-task processing-related information in a unified embedding space.

The interaction between the visual explicit-to-implicit transfer module and the language-guided global prompt is implemented in a cascaded manner. Specifically, the task-based extraction network first encodes visual information into intermediate representations. Concurrently, the linguistic information of low-level tasks is embedded in the global prompt matrix that conditions the intermediate representations through semantic modulation. Their multiplication is realized at the representation level with the implicit prompt matrix, ensuring that visual representations and prior linguistic information jointly contribute to the final prediction. In this way, these modules integrate pixel-level visual cues with high-level semantic priors in a unified representation space, enabling both task-agnostic and task-specific adaptation properties for controllable feature learning.

\subsection{Visual Explicit-to-Implicit Transfer}
\label{sec:3.2}
Although each visual image pair $I_{IN}'$ already contains unique explicit task-based information, the raw images do not contain sufficient task-related details. Especially in one-shot scenarios, it is necessary for a generalist model to extract stronger implicit information, which can guide the generalization process.

Therefore, we propose an explicit-to-implicit transfer module to benefit one-shot task generalization with rich implicit task-related information. Specifically, we utilize a DenseNet \citep{densenet} structure to extract this implicit context of task-related information due to its widespread adoption for feature capture. Previous work \citep{shallow} shows that shallow convolutional networks can capture efficient information on image generation tasks, so we choose a simplified DenseNet-18. This ensures that our explicit-to-implicit transfer module is both effective in implicit task-related learning.

Given a visual image pair input $I_{IN}'$, it is passed through the explicit-to-implicit transfer module. Thus we obtain a feature extraction matrix $ M_{F}$ as: 
\begin{equation}
    M_{F} = \mathcal{D}(I_{IN}')
\end{equation}
where $M_{F} \in \mathbb{R}^{p \times q}$, and $\mathcal{D}$ denotes the explicit-to-implicit task based extraction network. Through this transformation, explicit visual patterns are converted into high-level representations that act as an intermediate bridge to the following stage.

\subsection{Language-Guided Global Prompt}

While implicit visual representations tend to encode task-specific characteristics, implicit textural information provides task-agnostic cues that are also essential for one-shot learning. To further enhance one-shot performance, we present a language-guided prompt extraction strategy. The strategy extracts implicit textural information using a simple BERT-base-uncased \citep{devlin2019bert} tokenizer. For task name $L_{0}$ (\textit{e.g.}, ``dehazing''), we can obtain a language guided vector $V_L$ by:
\begin{equation}
   V_L = E(L_{0})
\end{equation}
where $E$ is the tokenizer module.

We also notice that, unlike input-dependent visual implicit extractions, task-related language information can also be performed with a global awareness. If we acquire a number ($n$) of textural names $L = \{L_{0}, L_{1}, ..., L_{n}\}$ from most low-level tasks, we can build a raw global prompt matrix $M_{L} \in \mathbb{R}^{n \times q}$:
\begin{equation}
    M_{L} = \textrm{concat}(\{E(L_{0}), E(L_{1}), ..., E(L_{n})\})
\end{equation}
With the additional global awareness, the combination of various language information encompasses a higher-level view of different tasks. In other words, $M_{L}$ captures the textual relationships among most low-level tasks within a latent space. To align with other modules, we also adjust the dimension of the raw matrix through a linear projection layer to receive the final language-guided global prompt matrix $M_{G}$:
\begin{equation}
        M_{G} = \textrm{linear}(M_{L}) \in \mathbb{R}^{q \times q}
\end{equation}

\subsection{Implicit Information Combination}
After introducing both implicit visual representations and language-guided global prompts, the next challenge is effectively integrating heterogeneous implicit cues into a cohesive representation. Intuitively, $M_{F}$ and $M_{G}$ represent implicit visual information and global textural information. The two matrices are able to be straightforwardly fed into the Painter decoder. However, a rough process that contains numerous distinct implicit information may mislead or overfit the model in practice.

To address the problem, we introduce an implicit learning matrix to further one-shot task generalization further. An implicit learning matrix $M_{I} \in \mathbb{R}^{p \times q}$ is set as a learnable parameter and designed with a selective mechanism by a multiplication operation.

Given the the global prompt matrix $M_{G} \in \mathbb{R}^{q \times q}$, implicit learning matrix $M_{I} \in \mathbb{R}^{q \times q}$, and the feature extraction matrix $M_{F} \in \mathbb{R}^{p \times q}$, the hidden product matrix $M_H \in \mathbb{R}^{p \times q}$ that comprises selected implicit visual and textural task-based information can be defined as:
\begin{equation}
    M_H = M_{F} \times (M_{G} \times M_{I}) 
\end{equation}

Due to the complexity of $M_H$, we deem it to be insufficient to be directly applied to the generalist model. To acquire richer dimensional prompts and unify feature space embeddings, we also leverage a frozen CLIP \citep{clip} model $\mathcal{C}$ to synthetically deal with all prompted components, including the visual image pair $I_{IN}'$, the hidden product matrix $M_H$, and the text prompt describing the task $P_T$.

Thinking of the base structure of CLIP, $\mathcal{C} = \mathcal{C}_{I} + \mathcal{C}_{T}$, where $\mathcal{C}_{I}$ and $\mathcal{C}_{T}$ represent CLIP image encoder and CLIP text encoder. The visual image pair $I_{IN}'$ is processed by the operation of patchify $\mathcal{O}_{P1}$, projection $\mathcal{O}_{P2}$, and adding positional encoding $\mathcal{O}_{P3}$.
\begin{equation}
I_{IN}'' = \mathcal{O}_{P1, P2, P3}(I_{IN}')
\end{equation}
Afterward, $I_{IN}''$ is combined with the hidden product matrix $M_H$ to the CLIP image encoder $\mathcal{C}_I$. On the textural part, the concise task-descriptive prompt $P_T$ (\textit{e.g.}, `This is a photo of [TASK NAME] task') is input into the CLIP text encoder $\mathcal{C}_T$. The coupling path to obtain the final prompt $P$ can be formalized as:
\begin{equation}
    P =\textrm{concat}(\mathcal{C}_{I}(\textrm{concat}(M_H, \mathcal{O}_{P1, P2, P3}(I_{IN}'))), \mathcal{C}_{T}(P_T))
\end{equation}

\begin{figure}[!t]
\centering
\includegraphics[width=1.0\columnwidth]{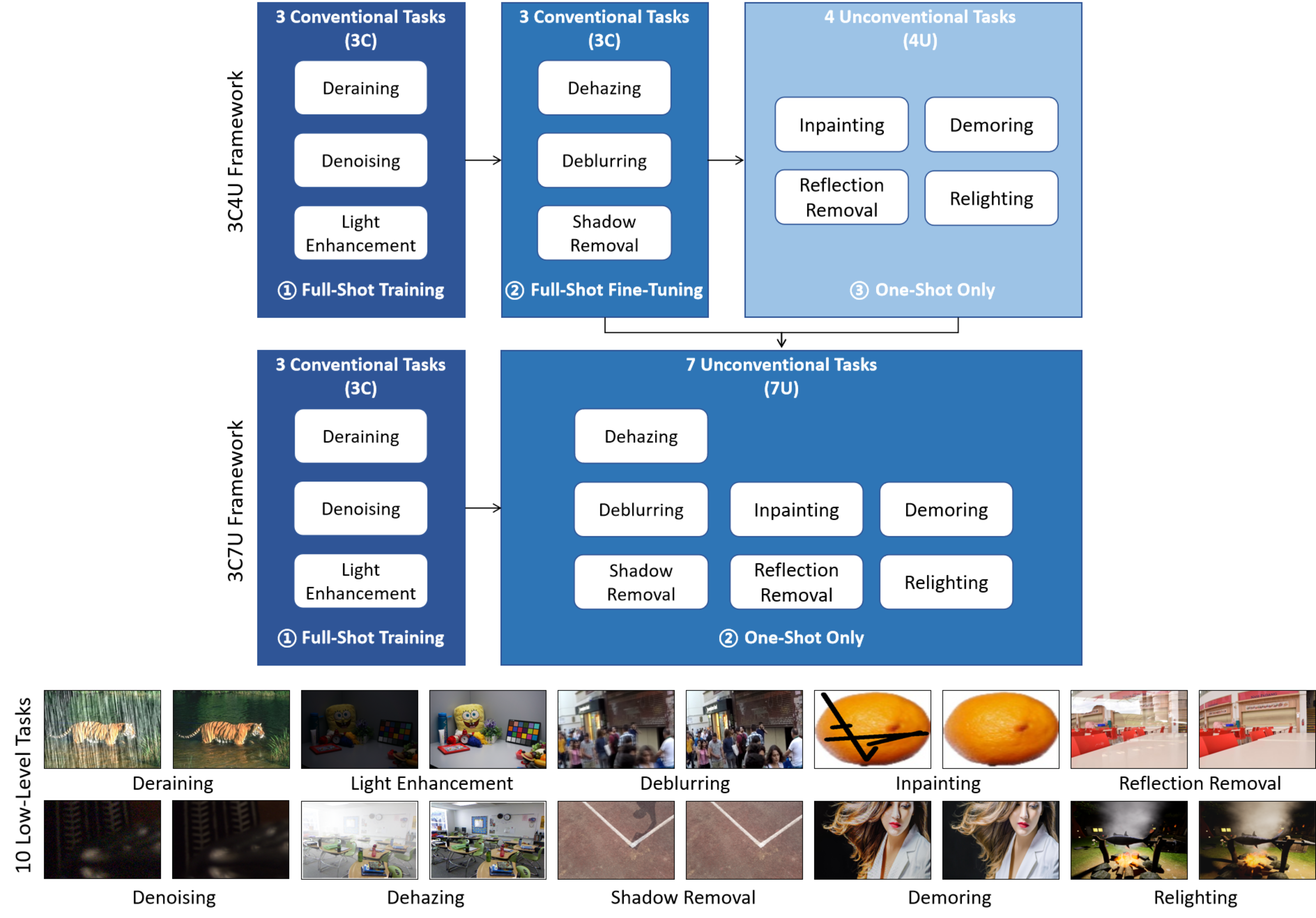}
\caption{3C4U Generalization Evaluation Framework. The framework consists of three stages: full-shot training, full-shot fine-tuning, and one-shot only.}
\label{fig:evalframework}
\end{figure}

\subsection{3C4U/3C7U Evaluation Framework}

Previous task generalization evaluation frameworks \citep{taskonomy} primarily focus on high-level vision tasks such as image segmentation, depth estimation, and object detection. To the best of our knowledge, no open-sourced and straightforward evaluation framework currently covers a sufficient number of low-level vision tasks and a wide range of task generalization types. As shown in Figure \ref{fig:evalframework}, we propose a data-driven evaluation framework termed the C/U assessment, which includes two representative configurations. The 3C4U setting (three conventional and four unconventional tasks) is used to retrain existing models, and the 3C7U setting (three conventional and seven unconventional tasks) is designed for training models from scratch. 

We categorize low-level vision tasks that frequently appear in studies \citep{jiang2025survey} of low-level vision as conventional tasks, such as draining, denoising, light enhancement, dehazing, deblurring, and shadow removal, whereas relatively niche or less commonly investigated tasks \citep{su2022survey} are defined as unconventional, such as inpainting, relighting, demoiring, and
reflection removal. Conventional tasks mainly include the learning of generic restoration priors and basic degradation-inversion mappings, while unconventional tasks challenge the learned representations by introducing more complex scene interactions and less frequently observed degradation patterns. Building upon this distinction, conventional tasks are mainly used to train the model to acquire fundamental low-level capabilities through a compact set of widely studied tasks, while unconventional tasks serve as the complementary indicators to assess the ability of the model to generalize to less common and more challenging low-level scenarios. This framework establishes a more comprehensive quantification standard for low-level generalist models, enabling a systematic evaluation of model generalization under heterogeneous low-level vision conditions.

In practical evaluation, two scenarios have to be considered: \textbf{Scenario 1}: From the perspective of training efficiency, utilizing existing generalist models is the optimal choice. Since most generalist models are primarily trained on high-level tasks with fewer low-level tasks, initial fine-tuning on more low-level tasks is necessary. \textbf{Scenario 2}: Regarding generalist models solely in low-level domain, directly training from scratch on some low-level tasks is another situation. Based on the above scenarios, the 3C4U evaluation framework consists of three/two stages:

(1) Initial full-shot training: The original vision generalist is trained on three predefined tasks (\textit{i.e.}, draining, denoising, light enhancement) to ensure a foundational understanding of low-level vision tasks. 
(2) Evaluation with full-shot fine-tuning: The generalist model is retuned with available input-ground truth of conventional tasks (\textit{i.e.}, dehazing, deblurring, and shadow removal) to assess the basic task generalization ability on low-level vision tasks. 
(3) Task generalization on one-shot only: Unconventional tasks (\textit{i.e.}, inpainting, relighting, demoiring, and reflection removal) are used for further rigorous evaluation of generalization capabilities in one-shot where only one input-output image pair is available for each task. For scenario 2, stage 2 and 3 are merged into the step of task generalization on one-shot only: Unconventional tasks (\textit{i.e.}, dehazing, deblurring, shadow removal, inpainting, relighting, demoiring, and reflection removal) are used for further rigorous evaluation of generalization capabilities in one-shot. We designate this minor distinction as 3C7U.

\section{Experiment}
\label{sec:experiments}
\subsection{Implementation Details}
\begin{figure}[tp]
\vspace{-7mm}
\centering
\includegraphics[height=0.6\textheight]{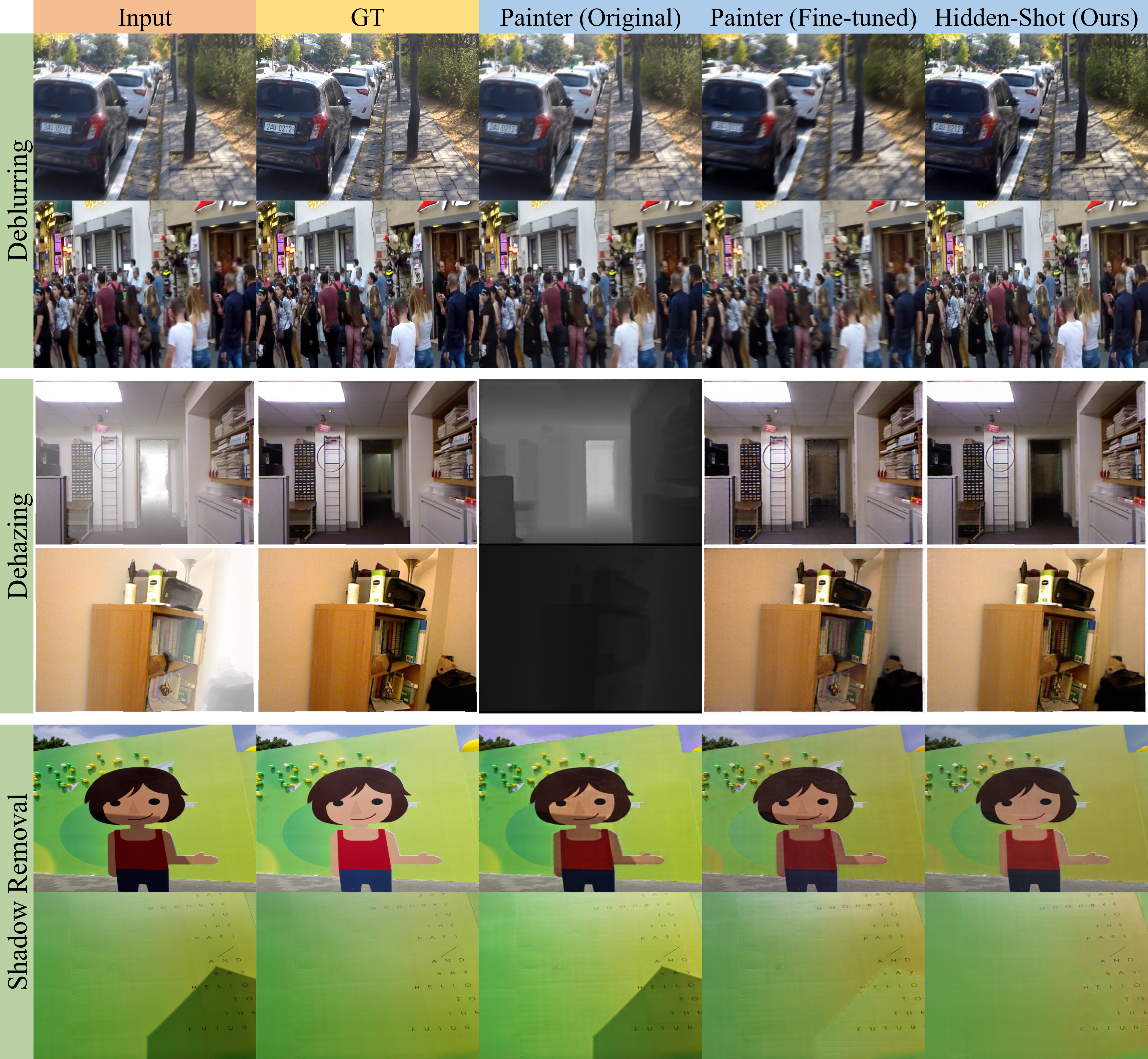}
\caption{Visual comparisons on three conventional tasks (3C). Hidden-Shot produces superior predictions than the baseline model for all three tasks.}
\label{fig:result1}
\end{figure}

\subsubsection{Datasets and Evaluation Metrics}
{\setlength{\tabcolsep}{2.3pt}
\renewcommand\arraystretch{0.7}
\begin{table}
  \caption{Comparisons of Hidden-shot with original Painter and fine-tuned Painter on 3C tasks (dehazing, deblurring, and shadow removal). Our method achieves competitive performance across the three tasks.}
  \label{tab:3c}
  \centering
  \begin{tabular}{llcccc}
    \toprule
    \multirow{2}{*}{{Task}} & \multirow{2}{*}{{Metric}} & \multicolumn{3}{c}{Method} \\
    \cmidrule(r){3-5}
    &  & Painter & Painter & Hidden-Shot \\
    & &(Original) & (Fine-tuned) & (Ours)\\
    \midrule
    \multirow{2}{*}{Dehazing}  
    & PSNR$\uparrow$ & 9.03  & 22.28  & 25.11  \\
    & SSIM$\uparrow$ & 0.399 & 0.849  & 0.884  \\
    \midrule
    \multirow{2}{*}{Deblurring}     
    & PSNR$\uparrow$ & 24.59 & 22.33  & 23.04  \\
    & SSIM$\uparrow$ & 0.742 & 0.733  & 0.743  \\
    \midrule
    Shadow  & PSNR$\uparrow$ & 19.63 & 25.55  & 26.22  \\
    Removal & SSIM$\uparrow$ & 0.743 & 0.775  & 0.800  \\
    \bottomrule
  \end{tabular}
\end{table}
}

\begin{figure*}[htbp]
\vspace{-5mm}
\centering
\includegraphics[width=\textwidth]{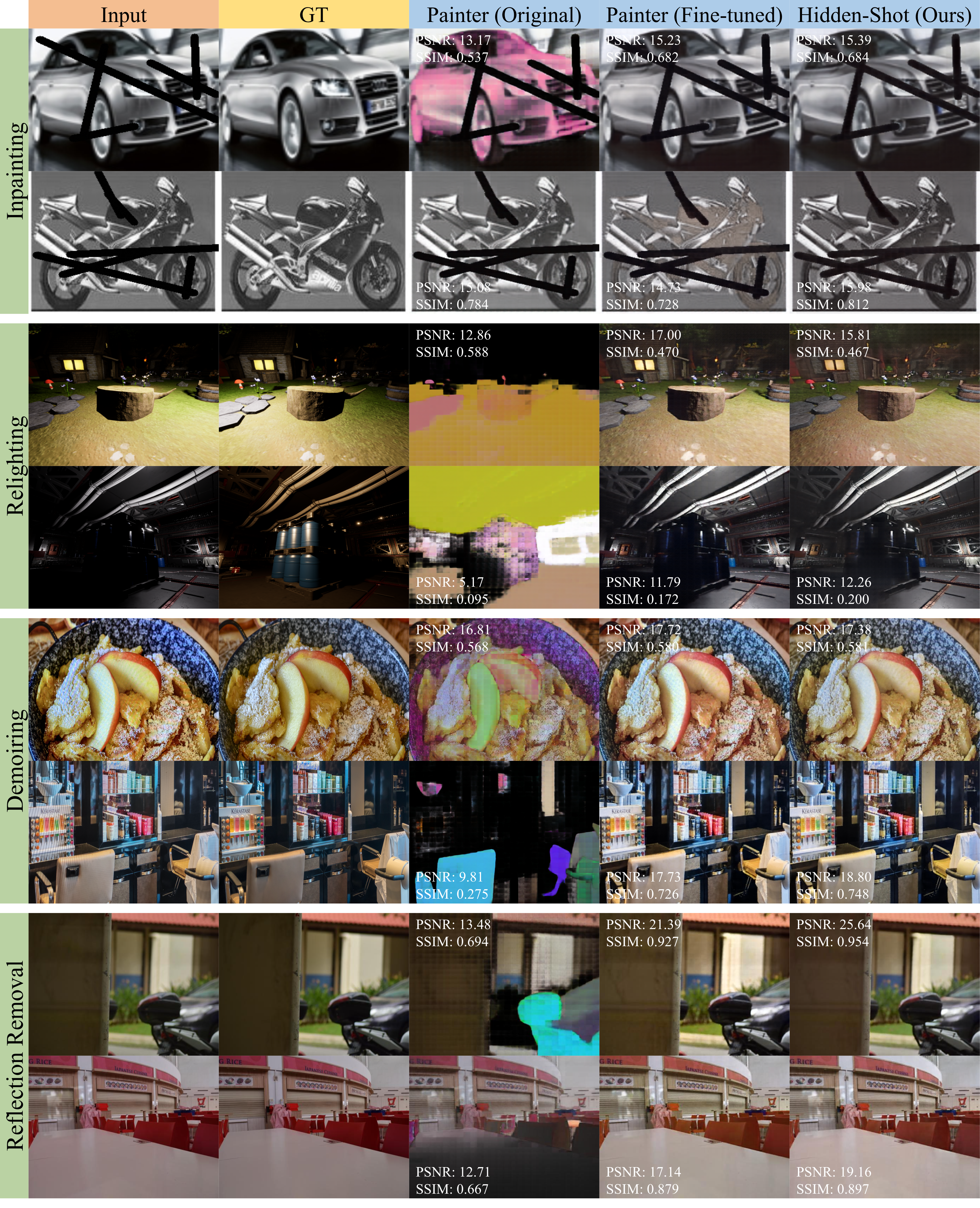}
\caption{Visual comparisons on four unconventional tasks (4U). Hidden-Shot consistently produces superior predictions compared to the baseline method. Meanwhile, quantitative metrics demonstrate the effectiveness of one-shot task generalization.}
\label{fig:result2}
\end{figure*}

For a fair and comprehensive examination, we conducted experiments on ten low-level tasks following proposed evaluation framework. All the datasets and corresponding low-level tasks are listed below: SIDD \citep{denoise} for denoising, LoL \citep{lol} for light enhancement, the merged deraining dataset \citep{derain} for deraining, D-Hazy \citep{dehazy} for dehazing, GOPRO Large \citep{deblur} for deblurring, and ISTD \citep{shadow} for shadow removal.
(2) Unconventional tasks: the generated inpainting dataset \citep{inpaint} for inpainting, VIDIT \citep{vidit} for relighting, 
UHDM \citep{uhdm} for demoiring, and SIR2 \citep{wan2017benchmarking} for reflection removal. The detailed description of each dataset is provided in \ref{add_data}. Referring to the previous study \citep{painter}, we report both results of peak signal-to-noise ratio (PSNR) and structural similarity index measure (SSIM).

\subsubsection{Modeling Details}
Painter \citep{painter} is selected as the baseline model. The proposed Hidden-Shot is injected into the baseline model. The structure is based on the illustration in Methodology. We include additional implementation details in \ref{imple}.

{\setlength{\tabcolsep}{2.3pt}
\renewcommand\arraystretch{0.7}
\begin{table}
  \caption{Comparisons of Hidden-shot with original Painter and fine-tuned Painter on 4U tasks (inpainting, relighting, demoiring, and reflection removal). The proposed method yields better generalization across the four unseen tasks.}
  \label{tab:4u}
  \centering
  \begin{tabular}{llcccc}
    \toprule
    \multirow{2}{*}{{Task}} & \multirow{2}{*}{{Metric}} & \multicolumn{3}{c}{Method} \\
    \cmidrule(r){3-5}
    &  & Painter & Painter & Hidden-Shot \\
    & &(Original) & (Fine-tuned) & (Ours)\\
    \midrule
    \multirow{2}{*}{Inpainting}        
    & PSNR$\uparrow$ & 15.74  & 15.06  & 15.07  \\
    & SSIM$\uparrow$ & 0.743  & 0.759  & 0.765  \\
    \midrule
    \multirow{2}{*}{Relighting}        
    & PSNR$\uparrow$ & 14.34  & 14.02  & 14.43  \\
    & SSIM$\uparrow$ & 0.510  & 0.202  & 0.185  \\
    \midrule
    \multirow{2}{*}{Demoiring}         
    & PSNR$\uparrow$ & 14.71  & 16.56  & 16.35  \\
    & SSIM$\uparrow$ & 0.614  & 0.682  & 0.678  \\
    \midrule
    Reflection
    & PSNR$\uparrow$ & 19.07  & 19.22  & 20.02  \\
    Removal & SSIM$\uparrow$ & 0.764  & 0.802  & 0.819  \\
    \bottomrule
  \end{tabular}
\end{table}
}

\subsection{First-phase Analysis on 3C4U framework}
\subsubsection{Results on New Full-Shot Fine-Tuned Tasks}
We conduct experiments on three new low-level training tasks, including dehazing, deblurring, and shadow removal. We present results in Table \ref{tab:3c} and Figure \ref{fig:result1}. Our approach consistently outperforms the original Painter and fine-tuned Painter on the three tasks. We observe a great improvement of PSNR (16.08 dB and 2.83 dB) and SSIM (0.485 and 0.035) on dehazing task. And the two indicators also increase 6.59 dB, 0.67 dB and 0.057, 0.025 on shadow removal. In the context of new training tasks, our results suggest that Hidden-Shot facilitates the generalist model in acquiring fresh knowledge more effectively.

{\setlength{\tabcolsep}{6.9pt}
\begin{table}[t]
\renewcommand\arraystretch{0.7}
  \caption{Analysis of stability on three low-level tasks from the original Painter paper (3C tasks: deraining, denoising, and light enhancement).}
  \label{tab:init}
  \centering
  \begin{tabular}{llccc}
    \toprule
    \multirow{2}{*}{{Task}} & \multirow{2}{*}{{Metric}} & \multicolumn{2}{c}{Method} \\
    \cmidrule(r){3-4}
    &  & Painter  & Hidden-Shot \\
    &  & (Train from & (Train from \\
    &  & scratch) & scratch) \\
    \midrule
    \multirow{2}{*}{Deraining}        
    & PSNR$\uparrow$ & 23.79  & 25.82  \\
    & SSIM$\uparrow$ & 0.782  & 0.837  \\
    \midrule
    \multirow{2}{*}{Denoising}        
    & PSNR$\uparrow$ & 30.04  & 31.66  \\
    & SSIM$\uparrow$ & 0.906  & 0.917  \\
    \midrule
    Light
    & PSNR$\uparrow$ & 21.20  & 22.08  \\
    Enhancement & SSIM$\uparrow$ & 0.763  & 0.797  \\
    \bottomrule
  \end{tabular}
\end{table}
}
\subsubsection{Performance on One-Shot Tasks}
In this section, four test sets of unseen tasks are fed into the model to make direct predictions. We completely expose Hidden-Shot to four one-shot tasks without any training---inpainting, relighting, demoiring, and reflection removal (4U). Table \ref{tab:4u} and Figure \ref{fig:result2} exhibit the results of our model, original Painter, and fine-tuned Painter. On the four tasks, Hidden-Shot obtains average PSNRs of 15.07 dB, 14.43 dB, 16.35 dB, 20.02 dB and mean values of 0.765, 0.185, 0.678, 0.819 for SSIM, respectively. In detail, we can further summarize the metrics into two main points. P1: In the realm of these uncommon low-level tasks, our model demonstrates notable advancements over original Painter and fine-tuned Painter. P2: Particularly in tasks where the original Painter performs poorly (PSNR $<$ 20 dB), our approach achieves considerable improvement in margins. \ref{add_vis} contains more abundant visualization results beyond Figure \ref{fig:result1} and Figure \ref{fig:result2}.

\subsection{Further Analysis on 3C7U framework}
\subsubsection{Results on Original Low-Level Tasks}
\label{sec:3.21}
In order to show our prompt framework can maintain the original low-level capabilities, we first assess the stability on three low-level tasks, \textit{i.e.}, deraining, denoising, and light enhancement (3C) from the original paper of Painter. Based on the fact that Hidden-Shot focuses solely on low-level tasks, we train from scratch on the same three datasets with the original weight of ViT-MAE \citep{vit-mae}. In Table \ref{tab:init}, the experiments indicate promotions of (2.03 dB, 0.055), (1.62 dB, 0.011), and (0.88 dB and 0.034) on PSNR and SSIM, respectively. Inspecting the original Painter paper, the training process also includes five high-level tasks that potentially influence low-level tasks, so we discard them. Thus our comparisons are strictly controlled over variables (only with low-level tasks during training). The superior performance demonstrates that our method retains the ability to handle the original three low-level tasks.

\subsubsection{Results on One-Shot Tasks}

In Table \ref{tab:4}, we conduct further experiments to evaluate the one-shot task generalization both on Hidden-Shot and Painter model. We employ models trained on the three tasks (3C) from the above subsection directly for inference. Table \ref{tab:4} demonstrates the one-shot performance on seven previously unseen tasks (7U), including dehazing, deblurring, shadow removal, inpainting, relighting, demoring, and reflection removal. Hidden-Shot consistently outperforms Painter in 13 metrics (7 PSNRs and 6 SSIMs). Precisely, our method achieves higher performance gains of (1.03 dB, 0.031), (1.87 dB, 0.033), (1.35 dB, 0.068), (1.23 dB, 0.033), (0.57 dB), (0.12 dB, 0.014), and (0.13 dB, 0.011) in PSNR and SSIM, respectively. These findings underscore the superior one-shot generalization capability of Hidden-Shot on a variety of unseen tasks.
% Hidden-Shot outperforms Painter in 13 metrics (7 PSNRs and 6 SSIMs).

\begin{table}
  \caption{Comparisons of one-shot task generalization between Hidden-Shot and Painter model across seven unseen tasks (7U). Both models were only trained on three low-level tasks (deraining, denoising, and light enhancement) outlined in the original Painter paper. Hidden-Shot outperforms Painter in 13 metrics (7 PSNRs and 6 SSIMs).}
  \label{tab:4}
  \centering
  \setlength{\tabcolsep}{7.1pt}  \renewcommand{\arraystretch}{0.7}
  
  \begin{tabular}{llccc}
    \toprule
    \multirow{2}{*}{{Task}} & \multirow{2}{*}{{Metric}} & \multicolumn{2}{c}{Method} \\
    \cmidrule(r){3-4}
    &  & Painter  & Hidden-Shot \\
    &  & (Train from & (Train from \\
    &  & scratch) & scratch) \\
    \midrule
    \multirow{2}{*}{Dehazing}         
    & PSNR$\uparrow$  & 10.23  & 11.26 \\
    & SSIM$\uparrow$  & 0.597  & 0.628 \\
    \midrule
    \multirow{2}{*}{Deblurring}       
    & PSNR$\uparrow$  & 21.10  & 22.97 \\
    & SSIM$\uparrow$  & 0.689  & 0.722 \\
    \midrule
    Shadow   
    & PSNR$\uparrow$  & 16.70  & 18.05 \\
    Removal & SSIM$\uparrow$  & 0.629  & 0.697 \\
    \midrule
    \multirow{2}{*}{Inpainting}       
    & PSNR$\uparrow$  & 14.00  & 15.23 \\
    & SSIM$\uparrow$  & 0.703  & 0.736 \\
    \midrule
    \multirow{2}{*}{Relighting}       
    & PSNR$\uparrow$  & 10.94  & 11.51 \\
    & SSIM$\uparrow$  & 0.174  & 0.157 \\
    \midrule
    \multirow{2}{*}{Demoiring}        
    & PSNR$\uparrow$  & 16.63  & 16.75 \\
    & SSIM$\uparrow$  & 0.668  & 0.682 \\
    \midrule
    Reflection
    & PSNR$\uparrow$  & 18.85  & 18.98 \\
    Removal & SSIM$\uparrow$  & 0.778  & 0.789 \\
    \bottomrule
  \end{tabular}
\end{table}

\subsection{Performance Analysis with Diffusion-Based Models}

As shown in Table \ref{tab:5}, Hidden-Shot demonstrates competitive performance across the evaluated low-level tasks. With either the 3C4U or 3C7U setting, Hidden-Shot achieves higher PSNR and SSIM than Prompt Diffusion and InstructDiffusion (with official pretrained weights) on most tasks. Relative to prompt- or instruction-based diffusion models, the inclusion of implicit visual features consistently yields promising VICL results. These results suggest that incorporating the implicit prompt mechanism contributes to stable generalizability and adaptability.

\begin{table*}[t]
  \centering
  \scriptsize
  \setlength{\tabcolsep}{3pt}
  \renewcommand{\arraystretch}{0.65}
  \caption{Comparison of Hidden-Shot with Prompt Diffusion and InstructDiffusion on the ten tasks.}
  \label{tab:5}
  \resizebox{\linewidth}{!}{
  \begin{tabular}{llcccc}
    \toprule
    \multirow{2}{*}{Task} & 
    \multirow{2}{*}{Metric} & 
    \multicolumn{4}{c}{Method} \\
    \cmidrule(r){3-6}
    & & Prompt Diffusion & InstructDiffusion & Hidden-Shot (3C4U) & Hidden-Shot (3C7U) \\
    \midrule

    \multirow{2}{*}{Deraining}        
    & PSNR$\uparrow$ & 10.23 & 15.39 & --    & 25.82 \\
    & SSIM$\uparrow$ & 0.096 & 0.433 & --    & 0.837 \\
    \midrule

    \multirow{2}{*}{Denoising}        
    & PSNR$\uparrow$ & 9.35  & 24.23 & --    & 31.66 \\
    & SSIM$\uparrow$ & 0.186 & 0.550 & --    & 0.917 \\
    \midrule

    \multirow{2}{*}{Light Enhancement}        
    & PSNR$\uparrow$ & 8.38  & 16.74 & --    & 22.08 \\
    & SSIM$\uparrow$ & 0.135 & 0.571 & --    & 0.797 \\
    \midrule

    \multirow{2}{*}{Dehazing}         
    & PSNR$\uparrow$ & 8.33  & 12.13 & 25.11 & 11.26 \\
    & SSIM$\uparrow$ & 0.150 & 0.558 & 0.884 & 0.628 \\
    \midrule

    \multirow{2}{*}{Deblurring}       
    & PSNR$\uparrow$ & 8.88  & 13.81 & 23.04 & 22.97 \\
    & SSIM$\uparrow$ & 0.161 & 0.369 & 0.743 & 0.722 \\
    \midrule

    \multirow{2}{*}{Shadow Removal}   
    & PSNR$\uparrow$ & 11.91 & 19.99 & 26.22 & 18.05 \\
    & SSIM$\uparrow$ & 0.187 & 0.508 & 0.800 & 0.697 \\
    \midrule

    \multirow{2}{*}{Inpainting}       
    & PSNR$\uparrow$ & 9.09  & 12.80 & 15.07 & 15.23 \\
    & SSIM$\uparrow$ & 0.303 & 0.465 & 0.765 & 0.736 \\
    \midrule

    \multirow{2}{*}{Relighting}       
    & PSNR$\uparrow$ & 6.78  & 9.27  & 14.43 & 11.51 \\
    & SSIM$\uparrow$ & 0.078 & 0.116 & 0.0.185 & 0.157 \\
    \midrule

    \multirow{2}{*}{Demoiring}        
    & PSNR$\uparrow$ & 6.48  & 12.21 & 16.35 & 16.75 \\
    & SSIM$\uparrow$ & 0.235 & 0.520 & 0.678 & 0.682 \\
    \midrule

    \multirow{2}{*}{Reflection Removal}
    & PSNR$\uparrow$ & 8.71  & 7.93  & 20.02 & 18.98 \\
    & SSIM$\uparrow$ & 0.177 & 0.358 & 0.819 & 0.789 \\
    \bottomrule
  \end{tabular}
  }
\end{table*}

\subsection{Ablation Studies}
\subsubsection{The Depth of Explicit-to-implicit Transfer Network}

Ablation experiments are performed on one hard task (reflection removal) with diverse testing samples, which also pose greater challenges in generalization. As shown in Table \ref{tab:ablation}, we investigate the influence of backbones varied at different depths, where the implicit learning matrix and global prompt matrix are treated as fixed components.

DenseNet-18 is set up as our reference baseline with increasing layer depths, which encompass DenseNet-121, DenseNet-169, and DenseNet-201. Comparisons across different DenseNet backbones show that increasing network depth does not yield improvements in PSNR or SSIM, suggesting that deeper networks provide limited additional benefit and may even lead to reduced efficiency. Due to the differences in parameter scale among DenseNet-18 and deeper backbones, the results reveal that employing a shallow backbone such as DenseNet-18 for feature extraction is an appropriate choice.

\subsubsection{The Effect of Global Prompt Matrix}

To explore and investigate the impact of the lightweight global prompt, an ablation experiment is executed to assess its contribution on the same DenseNet-18 extractor. Table \ref{tab:ablation} also shows the consequence of removing the global prompt matrix within the 3C4U evaluation framework. The PSNR values drop after removing the global matrix in both tasks. This drop indicates that the global prompt provides essential task-agnostic priors. Without it, the model relies only on local one-shot cues, leading to weaker contextual guidance and reduced consistency in the reconstructed outputs.

The results indicate that introducing subtle global prompts during training yields consistent performance gains in one-shot prediction, thereby enhancing the effectiveness of the implicit learning matrix. In terms of the implicit learning path is theoretically a black box, the designed global vectors make the process interpretable to some extent.
\begin{table}
  \caption{Ablation study on the depth of explicit-to-implicit transfer backbone and the impact of global prompt matrix.}
  \label{tab:ablation}
  \centering
  \setlength{\tabcolsep}{6pt}  \renewcommand{\arraystretch}{0.7}
  \begin{tabular}{lcc}
    \toprule
    \multirow{2}{*}{{Method}} & 
    \multicolumn{2}{c}{Reflection Remoal} \\
    \cmidrule(r){2-3}
    & PSNR$\uparrow$ & SSIM$\uparrow$ \\
    \midrule
    Hidden-Shot     & 20.02  & 0.819  \\
    DenseNet121     & 19.73  & 0.813  \\
    DenseNet169     & 19.75  & 0.814  \\
    DenseNet201     & 19.70  & 0.812  \\
    Non-Global      & 19.65  & 0.812  \\
    \bottomrule
  \end{tabular}
\end{table}

\subsection{Computational Efficiency}

To validate the lightweight and computationally efficient design of Hidden-Shot, we compare its model parameters, FLOPs, and actual inference latency with the competing methods mentioned before. Table \ref{tab:efficiency} mainly presents the results averaged across all tasks, while Table \ref{tab:efficiency_per_task} reports the results for each low-level task. Each individual task contains 100 randomly sampled images. The model incurs only limited additional computational cost over the underlying Painter framework and does not introduce noticeable inference latency overhead. 
Notably, even under FP32 inference, Hidden-Shot achieves substantially higher inference efficiency than the diffusion-based Prompt Diffusion and InstructDiffusion, despite their use of FP16 precision. 
This comparison confirms that Hidden-Shot maintains efficient one-pass inference while effectively avoiding the considerable computational overhead.
\begin{table*}[t]
  \centering
  \scriptsize
  \setlength{\tabcolsep}{5pt}
  \renewcommand{\arraystretch}{0.9}
  \caption{Comparison of inference efficiency at a fixed \(448\times448\) input resolution.}
  \label{tab:efficiency}
  \resizebox{\linewidth}{!}{
  \begin{tabular}{llrrr}
    \toprule
    Method &Default Settings & Parameters (B)$\downarrow$ & FLOPs (T)$\downarrow$ & Latency (s)$\downarrow$ \\
    \midrule
    Painter            & FP32, one pass & 0.371 & 1.600  & 0.268 \\
    Hidden-Shot        & FP32, one pass & 0.540 & 1.683  & 0.282 \\
    Prompt-Diffusion   & FP16, 50 steps & 1.733 & 83.998 & 2.144 \\
    InstructDiffusion  & FP16, 50 steps & 1.066 & 92.405 & 5.363 \\
    \bottomrule
  \end{tabular}
  }
\end{table*}

\subsection{Discussion}

Our experimental results across the seven and ten datasets indicate a consistent advantage over state-of-the-art vision generalist models, as validated under the 3C4U and 3C7U frameworks. This performance suggests that the proposed Hidden-Shot strategy not only preserves stable performance on existing tasks but also enables strong one-shot generalization to previously unseen tasks, highlighting its potential as a scalable and versatile solution for generalist models.

We also extend our comparison to include two widely used diffusion-based baselines, InstructDiffusion and Prompt Diffusion, to position our method against recent advances in instruction-guided diffusion models. Both baselines were provided with strong and well-defined text prompts to ensure high-quality generations and fair experimental conditions. Across the evaluated tasks, our approach consistently achieved better performance under both seen and unseen task settings.

In addition, we observe that diffusion models sometimes generate content that diverges substantially from the original image when performing VICL, including shape and color distortions, drastic style alterations, or implausibilities that were not present initially \citep{kamali2025characterizing}. Even when very strict positive and negative text-prompt constraints are applied, such generative behavior is misaligned with the objectives of our study, which require controlled and faithful preservation of the input structure. Hence, we adopt a relatively conservative framework to conduct our investigation in this work, deliberately avoiding generative components that may introduce confounding factors.

While our study demonstrates the notable marginal benefits of the Hidden-Shot approach, several potential limitations should also be acknowledged. Firstly, due to the substantial diversity of tasks within the low-level vision domain and the constraints on experimental time, we simplified our research framework by including a representative set of ten low-level tasks. A broader range of task combinations and more extensive experiments can be explored. Secondly, although our method demonstrates strong generalization ability across a variety of low-level vision tasks, our experiments also reveal several failure cases. The model tends to struggle in more challenging scenarios such as complex image-editing tasks and fine-grained edge detection. These tasks often require highly precise structural manipulation or extremely detailed boundary reasoning \citep{meng2024instructgie, soria2023tiny}, which might exceed the level of task cues provided by our current prompting mechanism. Lastly, as visual learning continues to progress toward more adaptable and generalizable paradigms, methods such as contrastive learning, continual learning, and knowledge replay emerge as promising future extensions. For example, Cheng et al. \citep{cheng2024progressive} propose a contrastive learning-based framework that progressively enhances negative samples to improve robustness across multiple dehazing scenarios. Meanwhile, the generalization problem across multiple adverse weather conditions is investigated through a continual learning framework equipped with effective knowledge replay in a unified network structure \citep{cheng2024continual}. In future work, we will explore the integration of these emerging techniques into our framework.

\section{Conclusion}
\label{sec: Conclusion}
We propose a simple and effective implicit prompt framework, \textit{i.e.}, Hidden-Shot for one-shot on heterogeneous low-level vision tasks. Our novel method yields stable improvement in one-shot of unseen low-level tasks and exceeds original performance on seen tasks, without significantly altering the network structure of the generalist model itself. The creative strategy well solves the gap between implicit prompt generation and lightweight global guidance when merging a new task. We expect that our findings will stimulate additional inquiry aimed at enhancing our comprehension on the one-shot prowess of generalist models.

\bibliographystyle{elsarticle-num} 
\bibliography{ref}

% https://aaai.org/conference/aaai/aaai-25/aaai-25-reproducibility-checklist/

\clearpage

\newpage
\appendix
\setcounter{table}{0}
\renewcommand{\thetable}{A\arabic{table}}

\setcounter{figure}{0}
\renewcommand{\thefigure}{A\arabic{figure}}

\begin{center} 
\textbf{\Large Appendix}
\end{center}

We further demonstrate our method by providing the following supplemental materials:
\begin{itemize}
\item Appendix A: Additional Dataset Details.
\item Appendix B: Additional Modeling Details.
\item Appendix C: Additional Visualization Results.
\end{itemize}
\section{Additional Dataset Details}
\label{add_data}
We present additional details of the ten datasets used in our experiment, following our 3C4U evaluation framework.
\subsection{Conventional Task Datasets}

We follow Painter \citep{painter} to prepare SIDD \citep{denoise} for denoising (13712 training images, 4300 testing images), LoL \citep{lol} for light enhancement (485 training images, 15 testing images), and also the merged deraining dataset \citep{derain} for deraining (485 training images, 15 testing images).

D-HAZY \citep{dehazy} is a widely-used dehazing dataset. There are around 400 images of hazy scenes and their corresponding ground truth. The dataset is build upon synthesizing haze in real images selected from Middlebury and NYU Depth datasets. Since the official training/testing split is not provided, we split the dataset following the 70/30 rule with randomized selection.

GOPRO Large \citep{deblur} is a famous deblurring dataset consisting of 3,214 blurred images with 1,280×720 resolution. We follow the official split method which encompasses 2,103 training images and 1,111 testing images.

ISTD \citep{shadow} is a dataset for shadow removal, which contains 1870 triplets of image shadow images, shadow masks, and original images. We follow the official training/testing split. However, only shadow images and original images are utilized due to the structure of the model input (both Painter \citep{painter} and Ours).

D-HAZY, GOPRO Large, and ISTD also serve as unconventional task datasets in 3C7U framework.

\subsection{Unconventional Task Datasets}
For image inpainting, we generate a randomly masked inpainting dataset using a natural image dataset \citep{inpaint} that consists of 6895 images from 8 distinct classes. 

VIDIT \citep{vidit} is a well-known relighting dataset that contains 390 different scenes. The dataset has five different types of color temperatures (2500K, 3500K, 4500K, 5500K, and 6500K) and different lighting directions (N, NE, E, SE, S, SW, W, NW). In our experiment, we report results on the official validation split (using both track 1 and track 2 and discarding depth maps) according to the CVPR 2021 NITRE challenge \citep{relight}.

UHDM \citep{uhdm} is a high-resolution demoiring dataset. In detail, 4,500 4K resolution training pairs and 500 standard 4K resolution validation pairs are split.

SIR2 \citep{wan2017benchmarking} is a widely-adopted reflection removal dataset containing 3 different types (postcard, solid object, and wild scene) of 500 triplets (images with reflection, reflection object, ground truth). The wild scene subset is considered the most difficult as it contains real-world objects of various complex reflectance. We use entire wild scene subsets as testing dataset.

\subsection{Input Size}
For fair comparisons, we use the input size of Painter and resize all the images and ground truth in visual prompted image pair $I_{IN}$ to 448$\times$448. Meanwhile, the input of Hidden-Shot (visual image pair $I_{IN}'$) is resized to 224$\times$224 for DenseNet-18.

% \begin{figure*}[tp]
% \centering
% \includegraphics[height=0.985\textheight]{Figure5_new.png}
% \caption{Additional visualization results following 3C4U evaluation framework.}
% \label{fig:add_vis_old}
% \end{figure*}

\begin{figure}[t]
  \centering
  \begin{minipage}{0.495\linewidth}
    \centering
    \includegraphics[width=\linewidth]{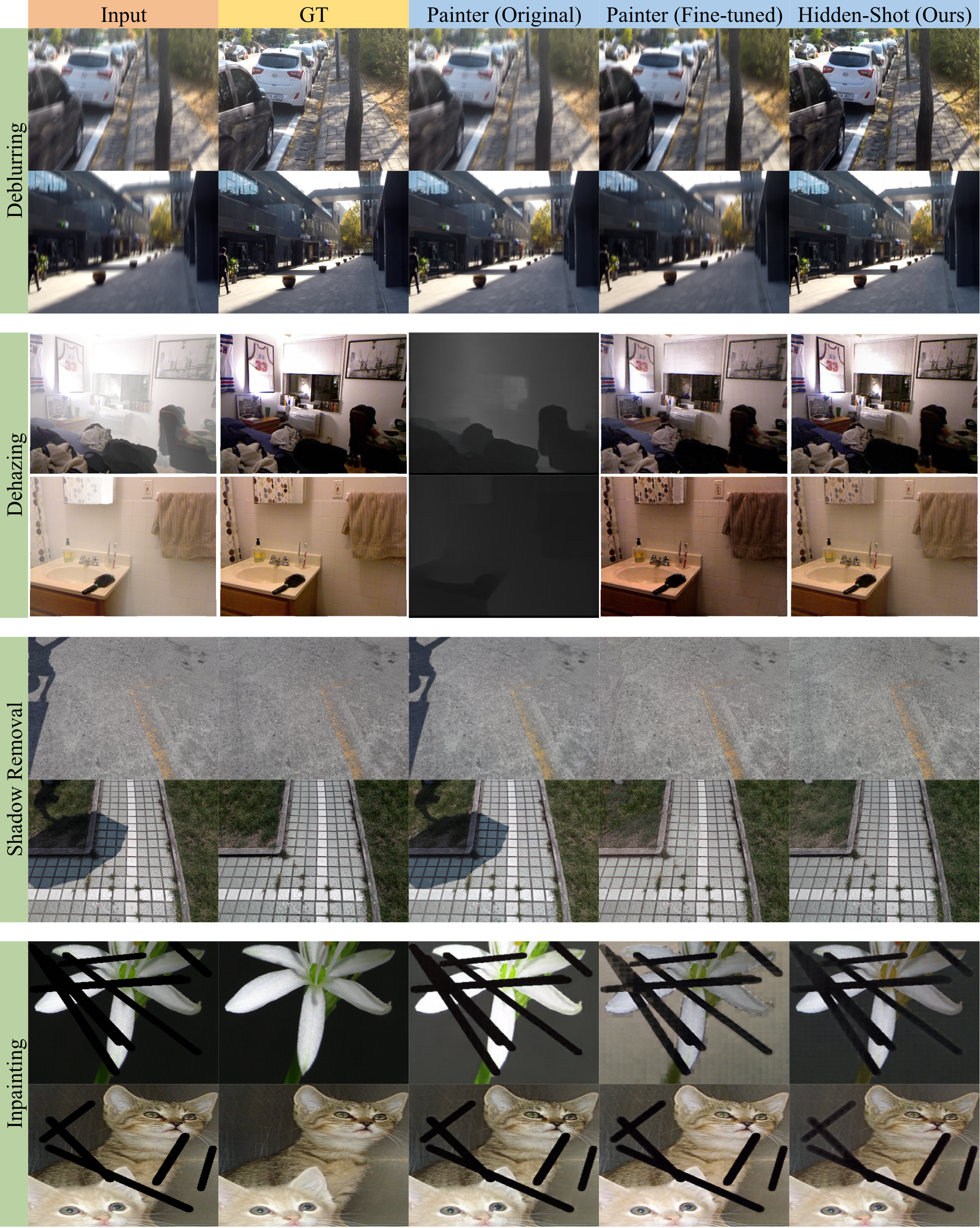}
  \end{minipage}
  \hfill
  \begin{minipage}{0.495\linewidth}
    \centering
    \includegraphics[width=\linewidth]{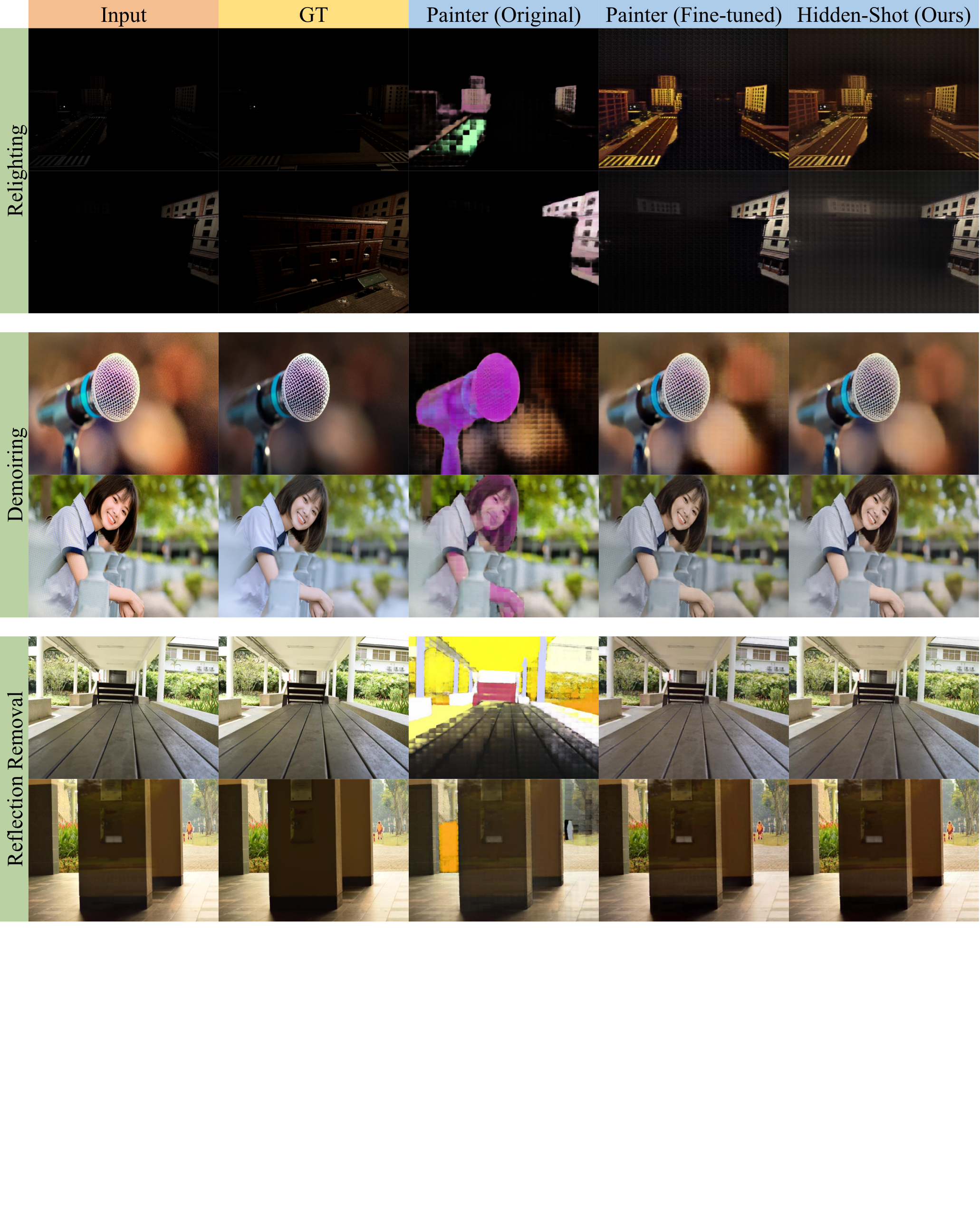}
  \end{minipage}
  \caption{Additional visualization results following 3C4U evaluation framework.}
  \label{fig:add_vis}
\end{figure}

\section{Module implementation details}
\label{imple}

Following the section of Methodology, we incorporate a DenseNet-18 \citep{densenet} model as the explicit-to-implicit transfer backbone, a BERT-base-uncased \citep{devlin2019bert} tokenizer (with sequences padded to a maximum length of 256) for implicit global textural prompt, and a pretrained frozen CLIP \citep{clip} model (ViT-B/32) for implicit information combination (implicit learning matrix: q = 256) and embedding space unification.

Original Painter, Instruct-Diffusion, and Prompt Diffusion all use the officially released weights for inference. For both our model and fine-tuned Painter, AdamW is utilized as the main optimizer with a weight decay of 0.1 and a layer decay of 1. Gradients are accumulated every 16 batches, with each batch comprising 4 sample pairs. The coefficients are also used for computing running averages of gradient, and their squares are fixed to 0.9 and 0.999.

In all subexperiments by 3C4U framework, training epochs are unified to 50 with initialized model weight from the original Painter \citep{painter} model. In all subexperiments following 3C7U framework, training epochs are arranged to 15 with initialized model weight from the original ViT-MAE \citep{vit-mae} model. The learning rate is set as 1e-3. We also include the original Painter under the 3C4U evaluation framework for further comparisons. All the models are written in PyTorch and trained on 4 NVIDIA A6000 GPUs (48GB each) from a Linux machine.

 In terms of computational efficiency, parameters are unique logical loaded parameters. FLOPs (T: teraFLOPs) count one multiply-add as two operations and cover the complete dynamic inference pipeline. Latency (B: billion) is the synchronized macro-task mean with batch size and per-process concurrency set to one; model loading and warm-up are excluded. Painter and Hidden-Shot are measured serially on one NVIDIA A100 GPU (80 GB). Painter and Hidden-Shot use the released FP32 single-forward inference, whereas Prompt Diffusion and InstructDiffusion use the default settings---FP16 with 50 inference steps. Table \ref{tab:efficiency_per_task} presents the metrics for each individual task.

\begin{table*}[t]
  \centering
  \scriptsize
  \setlength{\tabcolsep}{2.5pt}
  \renewcommand{\arraystretch}{0.95}
  \caption{Task-level comparison of inference efficiency at a fixed \(448\times448\) input resolution.}
  \label{tab:efficiency_per_task}
  % \resizebox{\linewidth}{!}{
  % \begin{tabular}{l*{10}{r}}
  %   \toprule
  %   & \multicolumn{10}{c}{Mean latency (s)$\downarrow$} \\
  %   \cmidrule(lr){2-11}
  %   Method
  %     & Deblurring & Dehazing & Demoiring & Denoising & Deraining & Light Enhancement & Shadow Removal & Reflection Removal & Relighting & Inpainting \\
  %   \midrule
  %   Painter
  %     & 0.283 & 0.212 & 0.644 & 0.206 & 0.212
  %     & 0.217 & 0.222 & 0.209 & 0.271 & 0.210 \\
  %   Hidden-Shot
  %     & 0.296 & 0.225 & 0.654 & 0.216 & 0.225
  %     & 0.225 & 0.237 & 0.221 & 0.288 & 0.227 \\
  %   Prompt-Diffusion
  %     & 2.273 & 2.078 & 2.453 & 2.142 & 2.073
  %     & 2.071 & 2.079 & 2.065 & 2.134 & 2.067 \\
  %   InstructDiffusion
  %     & 5.376 & 5.350 & 5.474 & 5.343 & 5.345
  %     & 5.346 & 5.348 & 5.343 & 5.367 & 5.343 \\
  %   \bottomrule
  % \end{tabular}
  % }

\resizebox{0.9\linewidth}{!}{
\scriptsize
  \begin{tabular}{l*{6}{c}}
    \toprule
    & \multicolumn{6}{c}{Mean Latency (s)$\downarrow$} \\
    \cmidrule(lr){2-7}
    Method
      & Deblurring & Dehazing & Demoiring & Denoising & Deraining & Relighting \\
    \midrule
    Painter
      & 0.283 & 0.212 & 0.644 & 0.206 & 0.212 & 0.271 \\
    Hidden-Shot
      & 0.296 & 0.225 & 0.654 & 0.216 & 0.225 & 0.288 \\
    Prompt-Diffusion
      & 2.273 & 2.078 & 2.453 & 2.142 & 2.073 & 2.134 \\
    InstructDiffusion
      & 5.376 & 5.350 & 5.474 & 5.343 & 5.345 & 5.367 \\
    \bottomrule
  \end{tabular}
}

  \vspace{0.2em}

\resizebox{0.9\linewidth}{!}{
  \scriptsize
  \begin{tabular}{l*{4}{c}}
    \toprule
    & \multicolumn{4}{c}{Mean Latency (s)$\downarrow$} \\
    \cmidrule(lr){2-5}
    Method
      & Light Enhancement & Shadow Removal & Reflection Removal & Inpainting \\
    \midrule
    Painter
      & 0.217 & 0.222 & 0.209 & 0.210 \\
    Hidden-Shot
      & 0.225 & 0.237 & 0.221 & 0.227 \\
    Prompt-Diffusion
      & 2.071 & 2.079 & 2.065 & 2.067 \\
    InstructDiffusion
      & 5.346 & 5.348 & 5.343 & 5.343 \\
    \bottomrule
  \end{tabular}
}

\end{table*}

\section{Additional Visualization Results}
\label{add_vis}
In this section, we showcase more abundant visual results in Figure~\ref{fig:add_vis} beyond Figure~\ref{fig:result1} and Figure~\ref{fig:result2}. These qualitative results indicate the robust performance of our method in task generalization surpassing the baseline model.

\end{document}